\documentclass[10pt,twocolumn,letterpaper]{article}

\usepackage[pagenumbers]{cvpr} %

\usepackage[dvipsnames,table]{xcolor}

\usepackage[accsupp]{axessibility}

\usepackage{graphicx}
\usepackage[export]{adjustbox} %
\usepackage{multirow}
\usepackage{makecell}
\usepackage{subcaption}
\newcommand{\thickhline}{\Xhline{3\arrayrulewidth}}
\newcolumntype{H}{>{\setbox0=\hbox\bgroup}c<{\egroup}@{}}
\newcommand{\tablestyle}[2]{\setlength{\tabcolsep}{#1}\renewcommand{\arraystretch}{#2}\centering\small}
\newcommand{\gain}[1]{{\color{blue}\scriptsize\textbf{#1}}}
\newcommand{\unsim}{\mathord{\sim}}  %

\newcommand{\myparagraph}[1]{\vspace{0pt}\noindent{\bf #1}}

\makeatletter
\@namedef{ver@everyshi.sty}{}
\makeatother

\usepackage{pgfplots}
\usepgfplotslibrary{fillbetween}
\usepackage{pgfkeys}
\def\addlegendimage{\csname pgfplots@addlegendimage\endcsname}

\usepackage[]{mdframed}

\definecolor{black}{rgb}{0.0, 0.0, 0.0}
\definecolor{blue}{rgb}{0.11764705882352941, 0.5647058823529412, 1.0}
\definecolor{green}{rgb}{0.66,0.82,0.56}
\definecolor{darkgreen}{rgb}{0.545, 0.749, 0.608}
\definecolor{darkergreen}{rgb}{0.384, 0.631, 0.576}
\definecolor{orange}{rgb}{0.9568627450980393, 0.3176470588235294, 0.11764705882352941}
\definecolor{Gray}{gray}{0.9}

\definecolor{GRed}{rgb}{0.8,0.0,0.0}
\definecolor{GBlue}{rgb}{0.2588,0.5216,0.9569}
\definecolor{GGreen}{rgb}{0,0.5922,0.5921}
\definecolor{GOrange}{rgb}{1,0.6706,0.2510}

\def\vc{{\mathbf{c}}}
\def\vs{{\mathbf{s}}}
\def\vu{{\mathbf{u}}}
\def\vv{{\mathbf{v}}}
\def\vx{{\mathbf{x}}}
\def\vy{{\mathbf{y}}}
\def\vz{{\mathbf{z}}}
\def\to{$\rightarrow$}

\usepackage{pifont}
\newcommand{\cmark}{\text{\ding{51}}}
\newcommand{\xmark}{\text{\ding{55}}}

\newcommand{\shortrefsec}[1]{\S\ref{sec:#1}}

\def\vs{\emph{vs}\xspace}
\def\n{\textbackslash n}
\def\hquad{\hspace{0.5em}}

\pgfplotsset{compat=1.15}

\definecolor{cvprblue}{rgb}{0.21,0.49,0.74}
\usepackage[pagebackref,breaklinks,colorlinks,citecolor=cvprblue]{hyperref}

\def\paperID{1502} %
\def\confName{CVPR}
\def\confYear{2024}

\title{
Distilling Vision-Language Models on Millions of Videos %
}

\author{
    Yue Zhao$^{1,2}$\thanks{Work done during an internship at Google Research.} \quad Long Zhao$^{1}$ \hquad Xingyi Zhou$^{1}$ \hquad Jialin Wu$^{1}$ \hquad Chun-Te Chu$^{1}$ \hquad
    Hui Miao$^{1}$ \hquad
    Florian Schroff$^{1}$ \\ Hartwig Adam$^{1}$ \quad
    Ting Liu$^{1}$ \quad Boqing Gong$^{1}$ \quad Philipp Kr\"ahenb\"uhl$^{2}$ \quad Liangzhe Yuan$^{1}$ \\
    $^{1}$Google Research \quad $^{2}$University of Texas, Austin \\
}

\begin{document}
\maketitle

\begin{abstract}
The recent advance in vision-language models is largely attributed to the abundance of image-text data.
We aim to replicate this success for video-language models, but there simply is not enough human-curated video-text data available.
We thus resort to fine-tuning a video-language model from a strong image-language baseline with synthesized instructional data.
The resulting video model by video-instruction-tuning (\textbf{VIIT}) is then used to auto-label millions of videos to generate high-quality captions.
We show the adapted video-language model performs well on a wide range of video-language benchmarks.
For instance, it surpasses the best prior result on open-ended NExT-QA by 2.8\%.
Besides, our model generates detailed descriptions for previously unseen videos,
which provide better textual supervision than existing methods.
Experiments show that a video-language dual-encoder model contrastively trained on these auto-generated captions is 3.8\% better than the strongest baseline that also leverages vision-language models.
Our best model outperforms state-of-the-art methods on MSR-VTT zero-shot text-to-video retrieval by 6\%.
As a side product, we generate the largest video caption dataset to date.

\end{abstract}
\section{Introduction}
\label{sec:intro}

Much progress in image understanding~\cite{zhou2022detic,openai2023gpt4v,yuan2021florence,dosovitskiy2021vit,wang2023beit3} is fueled by large-scale high-quality image-text datasets~\cite{kirillov2023segmentanything,chen2023pali,radford2021clip,schuhmann2022laion5b}.
Despite the wide availability on the Internet, annotating videos is nontrivial.
For images, humans construct most annotations within $15\unsim90$ seconds per instance~\cite{lin2014coco,kirillov2023segmentanything}.
For videos, the annotation time is $1\unsim2$ orders of magnitude higher: it takes about 70 hours to transcribe narratives for a 1-hour video~\cite{grauman2022ego4d,xu2016msrvtt} and 700 hours to provide a 1-hour video with instance-level annotations~\cite{darkhalil2022epic}.
There have been attempts to automate such a process by retrieving alt-text~\cite{bain2021webvid,nagrani2022videocc} or transcribing text from audio~\cite{miech2019howto100m,zellers2022merlot}.
However, alt-text can be irrelevant to the video content; audio transcription is often misaligned with the visual information~\cite{han2022tan}.
{Recent work~\cite{wang2023internvid} leverages existing image-based vision-language models (VLMs).
However, the resulting captions are often biased towards static scenes and lose videos' rich temporal information.
}

\begin{figure}[!t]
    \centering
    \includegraphics[width=\linewidth]{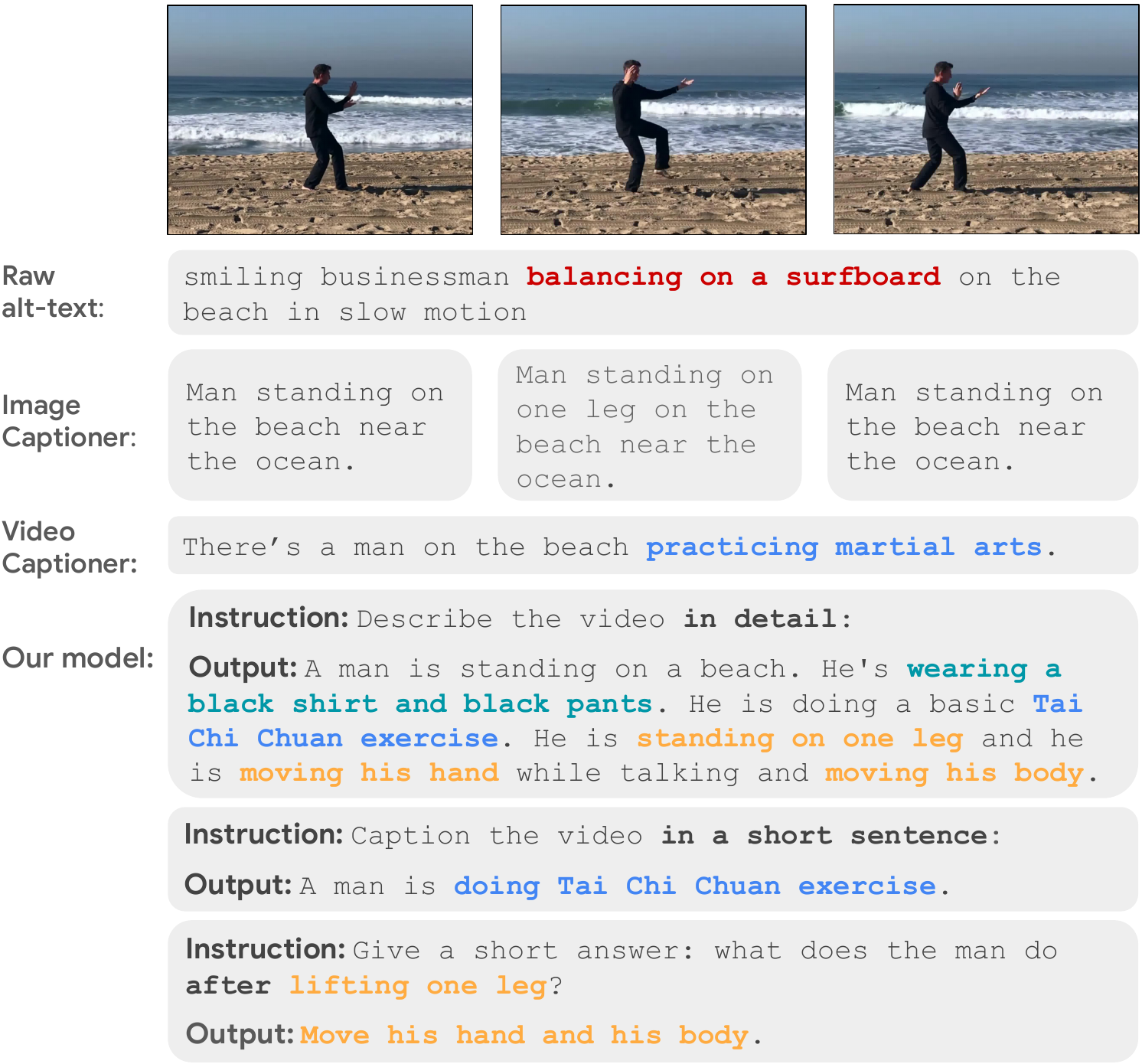}
    \vspace{-15pt}
    \caption{Our video-language model takes a video along with \textit{any} form of instruction as input and generates text according to the instruction.
    It generates textual descriptions with multiple granularities, including  {\color{GGreen}\textbf{static appearance}}, {\color{GBlue}\textbf{general action}}, and {\color{GOrange}\textbf{detailed body movements}}.
    In contrast,
    raw alt-text can be {\color{GRed}\textbf{erroneous}};
    image captioners fail to capture the action;
    video captioners prefer outputting short text.
    Our generated data trains a significantly better video-language dual-encoder model.
    Best viewed in color.
    }
    \label{fig:teaser}
    \vspace{-5pt}
\end{figure}

In this paper, we propose a simple yet effective approach to adapt an image-based VLM to video and then
create
high-quality pseudo-captions on millions of videos.
{As a VLM is generally composed of a visual encoder and a language model, we propose to adapt each component separately to better leverage the relatively small video-text corpora.}
We first fine-tune the visual encoder on video captioning data while keeping the language component frozen.
This adapts the model to dynamic scenes while retaining the diverse ability of the original language decoder.
We then fine-tune the language model on a small amount of instruction-following data and keep the visual encoder frozen.
This is to emphasize the temporal and causal reasoning ability beyond scene-level description.
The resulting video-language model sees both dynamic input and motion-focused output and is capable of generating high-quality pseudo-captions for million-scale web-scraped videos.

{Pseudo-captioning by the adapted VLM have the following advantages.
First, the captions are generally relevant to visual content because of the maximum likelihood objective during video-captioning training.
Second, our pseudo-captions preserve temporal information in videos better than frame-wise captions for videos~\cite{wang2023internvid,maaz2023videochatgpt}.
Third, the instruction-tuned video-language model generates textual descriptions with multiple granularities, including static appearance, general actions, and detailed body movements.
Finally, compared to human labeling, pseudo-captioning is more scalable.
For each video, the underlying language model can output multiple candidate captions in parallel in a single pass,
and the annotation cost
can be further improved given
advances in efficient inference techniques~\cite{leviathan2023speculativedecoding}.
}

We evaluate the resultant VLM on a wide range of video-language benchmarks, covering video question answering (QA) and captioning, and observe state-of-the-art zero-shot performance on all.
For instance, it attains a 29.5\% WUPS score on open-ended NExT-QA, 2.8\% better than Flamingo-80B while using only $\frac{1}{16}\times$ parameters.
We further use this adapted VLM to generate video descriptions on million-scale web-scraped videos.
Qualitatively, the generated descriptions are more specific and detailed than alt-text or image captions.
To evaluate the pseudo-captions quantitatively, we train a CLIP-style~\cite{radford2021clip} video-language dual-encoder model using the generated descriptions.
We observe a striking scaling effect on the performance with respect to the size of pseudo-captioned video data, which does not hold for alt-text alternatives.
{Our model also works better than the one trained on frame-wise captions followed by LLM summarization.}
Notably, the dual-encoder model trained on 17 million web-scraped video clips with our machine-generated descriptions achieves the state-of-the-art performance on popular video-text retrieval and video recognition benchmarks.
For instance, the model scores 48.4\% Recall@1 on MSR-VTT, 6\% higher than the best previously reported number.

\section{Related Work}
\label{sec:related}

\myparagraph{Synthetic data}
from simulators are useful to create new datasets or augment existing ones~\cite{de2022next} for vision tasks such as optical flow~\cite{dosovitskiy2015flownet}, semantic segmentation~\cite{richter2016playing}, and 3D vision~\cite{chang2015shapenet}.
LLM-generated text becomes a great source for language understanding~\cite{meng2022generating}.
For example, Vicuna~\cite{vicuna2023} fine-tunes LLaMA~\cite{touvron2023llama} on user-shared conversations from ShareGPT. 
In the context of vision-language understanding, generating high-quality synthetic captions for vision data by leveraging LLMs has been shown effective in improving multimodal datasets for VLMs~\cite{nguyen2023improving}.
VideoChatGPT~\cite{maaz2023videochatgpt} uses both human-assisted and semiautomatic annotation methods with BLIP-2~\cite{li2023blip2} and GPT-3.5 to generate high-quality video instruction data. InternVid~\cite{wang2023internvid} introduces a scalable approach to automatically construct a high-quality video-text dataset with BLIP-2 and Vicuna.
LLaVA~\cite{liu2023llava} incorporates instruction tuning to VLMs, which demonstrates impressive multi-modal chat abilities.
However, these methods either focus on image inputs or rely on image models to produce video captions, which fail to capture correct temporal information in videos.

\myparagraph{Vision-language models.} Utilizing image-text data for pre-training has become the default approach to tackle vision-language tasks. Recently,  VLMs based on image-text contrastive learning (\eg, CLIP~\cite{radford2021clip} and ALIGN~\cite{jia2021align}) attain strong results on zero-shot retrieval and classification tasks. Follow-up studies propose to add more pre-training objectives, such as captioning loss (\eg, CoCa~\cite{yu2022coca}), to enable VLMs to handle different downstream tasks (\eg, image captioning and visual QA). Parallel methods explore leveraging off-the-shelf pre-trained models and keep them frozen during training. They partially freeze either vision or language models (\eg, PaLI~\cite{chen2023pali,chen2023palix,chen2023pali3} and LiT~\cite{zhai2022lit}) or insert new layers between them (\eg, Flamingo~\cite{alayrac2022flamingo} and BLIP-2~\cite{li2023blip2}) so that the knowledge from frozen models can be transferred to vision and language tasks. Our work builds upon them and tackles video inputs, a more challenging modality involving temporal and causal reasoning of motion.

\myparagraph{Video-language models}
can be adapted from image-language models given that image-based foundation models are pre-trained on web-scale image data.
VideoCLIP~\cite{xu2021videoclip} leverages a pre-trained CLIP model~\cite{radford2021clip} as a frame-level feature extractor and fine-tunes video and text transformers on video datasets.
VideoCoCa~\cite{yan2022videococa} builds on CoCa~\cite{yu2022coca} and fine-tunes some temporal pooler layers to reason over time.
Another line of research focuses on parameter efficient tuning, which is first shown effective on language modeling~\cite{lester2021pet}.
AIM~\cite{yang2023aim} adapts pre-trained image models for efficient video understanding by freezing pre-trained weights and tuning a few lightweight adapters.
Furthermore, to solve more complex video-language tasks like captioning and QA, researchers leverage the powerful LLMs as a universal interface and adapt LLMs to consume visual tokens.
FrozenBiLM~\cite{yang2022frozenbilm} leverages a frozen bi-directional language model for video QA.
VideoChat~\cite{li2023videochat} and VideoChatGPT~\cite{maaz2023videochatgpt} propose a chatbot-like interface to analyze video input.
However, VideoChat only shows qualitative analysis while VideoChatGPT relies on a GPT-4 for quantitative evaluation, leading to inconsistency over time.
LaViLa~\cite{zhao2023lavila} develops a video-language model that densely narrates for a video.
However, training the narrator assumes videos to be partially annotated.
Our work takes a further step and shows that the adapted video-language model generalizes to million-scale \textit{unseen} videos.

\section{Preliminaries and Notations}
\label{sec:prelim}
We first describe preliminaries and, meanwhile, introduce some notations facilitating the presentation of our method. 

\myparagraph{Image-based VLMs}
take as input an image and a text sequence, which is often called a prompt~\cite{brown2020gpt3} or an instruction~\cite{wei2022flan}, and outputs another text sequence that follows the prompt.
Specifically, let $\vx\in\mathbb{R}^{H\times W\times 3}$ denote an input image with height $H$ and width $W$, $\vy=(s_1, \cdots, s_{L_i}) \in\{0, 1\}^{{L_i}\times|\mathbb{S}|}$ the instruction, and $\vz=(z_{1}, \cdots, z_{L_o})\in\{0, 1\}^{{L_o}\times|\mathbb{S}|}$ the output text that are tokenized~\cite{kudo2018sentencepiece} into sequences of discrete symbols.
Here $\mathbb{S}$ denotes the vocabulary set, and $L_i$ and $L_o$ are the sequence lengths of the instruction and output, respectively.

A typical VLM has a visual encoder $F_V$ and a language model $F_L$.
The visual encoder maps $\vx$ to $N$ visual tokens $\vx' = F_V(\vx)\in\mathbb{R}^{N\times C}$.
It is often instantiated by a pre-trained Convolutional Network~\cite{he2016resnet} or Vision Transformer~\cite{dosovitskiy2021vit} plus an optional projection module in the form of Q-Former~\cite{li2023blip2}, Resampler~\cite{alayrac2022flamingo}, or attentional pooler~\cite{yu2022coca}.
The language model projects an input instruction $\vy$ to text tokens $\vy'\in\mathbb{R}^{L_i \times C}$, concatenates them with the visual tokens, and emits a text sequence recursively $\tilde{z}_l = F_L(\vx',\vy', \vz_{<\ell)}$, where ${\vz_{<\ell}}=[\tilde{z}_0, \cdots, \tilde{z}_{l-1}]$ with $\tilde{z}_0$ being a special start-of-sentence token \texttt{<s>}.
$F_L$ can be either an encoder-decoder-style model~\cite{raffel2020t5,tay2023ul2}, or a decoder-only model~\cite{brown2020gpt3}.
In this paper, we train the VLM using a captioning loss,~\ie, the sum of the negative log-likelihood of the correct word at each step:
{\small
\begin{align}
    \mathcal{L}=-\sum_{\ell=1}^L p(z_\ell | \vx', \vy', \vz_{<\ell}).
    \label{eq:caption_loss}
\end{align}
}
The key to the recent success of VLMs is the abundance of paired image-text datasets $\{(\vx, \vc)\}$.
By setting $\vy=\varnothing$ or a fixed task prompt for captioning and $\vz=\vc$, we can easily scale up VLMs by training on billion-scale datasets~\cite{chen2023pali,schuhmann2022laion5b}. 

\myparagraph{Visual instruction tuning}
intends to enable VLMs to tackle tasks beyond image captioning~\cite{liu2023llava}.
In this case, $(\vy, \vz)$ can be a question-answer pair as in visual QA~\cite{goyal2017vqa2}, or more generally, any free-form instruction-answer pair. 
The paired instruction-answer data are typically transformed from a plain caption via few-shot prompting by a language model~\cite{brown2020gpt3,wang2023selfinstruct},~\ie $(\vy, \vz)=\mathrm{LLM}(\vc)$.

\begin{figure*}[!tb]
    \centering
    \includegraphics[width=0.75\linewidth]{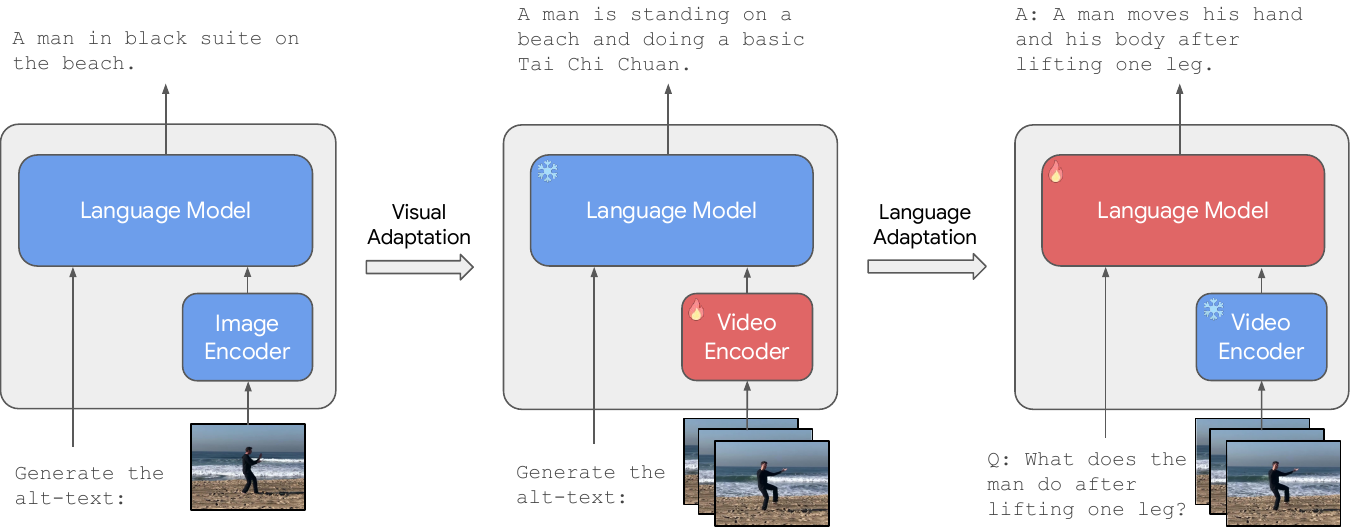}
    \vspace{-5pt}
    \caption{
    \textbf{Overview of adapting vision-language models to videos.}
    In the first stage of visual adaptation on sequences of video frames, we fine-tune the vision encoder while freezing the language model using a video dataset with captions.
    In the second stage of language adaptation, we freeze the vision encoder while fine-tuning the language model using a video dataset with instruction-following data,~\eg a question that requires temporal reasoning to answer in this example.
    }
    \label{fig:overview}
    \vspace{-5pt}
\end{figure*}

\myparagraph{Video-text datasets.}
One of the main challenges in training video-language models is the lack of video-text data.
The largest public video dataset with human-labeled textual descriptions is Spoken Moments in Time (S-MiT)~\cite{monfort2021smit}, which has $\unsim$500K videos.
Although the covered topics are diverse, the video durations are short ($2\unsim 3$ seconds), and the captions are brief.
The textual descriptions are transcribed from audio recordings with inevitable transcription errors.
The Video Localized Narratives (VidLN)~\cite{voigtlaender2023vidln} dataset captures more complex events for longer videos ($10\unsim30$ seconds), but it is $10\times$ smaller in the number of videos due to annotation cost.
Both lag in scale far behind existing image-text datasets,~\eg WebLI-10B and LAION-5B.
In the following section, we present an approach to leveraging these existing video-text datasets to efficiently adapt a pre-trained VLM from images to videos so that we can obtain high-quality pseudo-captions for millions of in-the-wild videos.
Experiments show our method yields competitive annotation quality and is more scalable than human annotation for videos.

\section{Method: Adapting VLMs to Videos}
\label{sec:method}
We adapt an image-language model to the video domain in two stages. In the first stage, we adapt the visual encoder while freezing the language component, allowing us to leverage relatively large video-text datasets whose text is unfortunately short and low-quality. In the second stage, we finetune the language encoder and freeze the other model components using a smaller video-text dataset whose text describes the video in detail and provides diversity. We empirically justify the advantage of this two-stage design, which is necessary given the video-text data's quality and size falling behind its image-text counterpart.

\subsection{Model}
Our video-language model takes a sequence of frames as visual input.
Let $\{\vx_1, \cdots, \vx_T\}$ denote the input video, where $T$ is the number of frames.
We pass each frame $\vx_t$ into the visual encoder $F_V$ and concatenate all output visual tokens, namely $\vx'=[F_V(\vx_1), \cdots, F_V(\vx_T)]\in \mathbb{R}^{TN\times C}$.
By doing so, we maintain the visual modeling capacity from the image-based models~\cite{chen2023pali3} and keep both computation and memory cost tractable ($O(TN^2)$ rather than $O(T^2N^2)$).
The language model then collects the visual tokens plus input instruction tokens and emits a text sequence.

\myparagraph{Model architecture.} We start with PaLI-3~\cite{chen2023pali3}, a state-of-the-art VLM trained on WebLI~\cite{chen2023pali} which has image-text data only.
The visual encoder is a ViT-G/14~\cite{zhai2022scalingvit} with 2B parameters.
The language model follows an encoder-decoder architecture based on UL-2~\cite{tay2023ul2} with 3B parameters.
We feed the adapted model with 8 frames at 2 FPS and resize the input resolution to $224\times 224$.

\subsection{Two-Stage Adaptation}
\label{sec:method:adaptation}
Due to the scarcity of paired video-text data, we propose to fine-tune the video-language model from the image-based baseline in two stages:
(1) visual adaptation, where we freeze the language component while fine-tuning the visual part with a relatively large video dataset with short captions; and (2) language adaptation, where we instruction-tune the language component while freezing the visual part with a smaller video dataset with detailed captions.

\myparagraph{Visual adaptation.}
In the stage of visual adaptation, we fine-tune $F_V$ while keeping $F_L$ frozen using a large video dataset with short captions $\{(\vx, \vc)\}$.
We optimize~\cref{eq:caption_loss} by setting $\vy$ to be a fixed task prompt for captioning (``\texttt{Generate the alt-text:}") and $\vz$ to be the caption.
On one hand, finetuning $F_V$ enables the visual encoder to focus more on scene dynamics rather than appearance.
On the other, freezing $F_L$ prevents the language model from possible collapse due to simple text and repetitive patterns.

\myparagraph{Language adaptation.}
In this stage, we fine-tune $F_L$ while keeping $F_V$ frozen using videos with instruction-following data generated as follows.
Given a video $\vx$ and its caption $\vc$, we first prompt an LLM to generate a question $\vy$ and the corresponding answer $\vz$ which is inferred from the original caption.
We optimize~\cref{eq:caption_loss} with the $(\vx, \vy, \vz)$ triplets.

The video-language model's temporal reasoning ability is highly dependent on the instruction-following data it trains on. 
To this end, we design prompts to encourage LLMs to generate \emph{causal} and \emph{temporal} questions inspired by how the NExT-QA dataset~\cite{xiao2021nextqa} is constructed.
Causal questions either explain the intention of an action that happens first or the cause of an action that occurs next.
It typically follows the form of ``Why did somebody do something?" or ``How did something happen?".
Temporal questions ask about the temporal ordering of multiple actions.
The temporally ordered actions can either happen on a single object or occur between multiple persons or objects.
We provide an example for illustration in \Cref{fig:instruction_examples} and more details in the supplementary materials.

\begin{figure*}[!tb]
\centering
\includegraphics[width=0.88\linewidth]{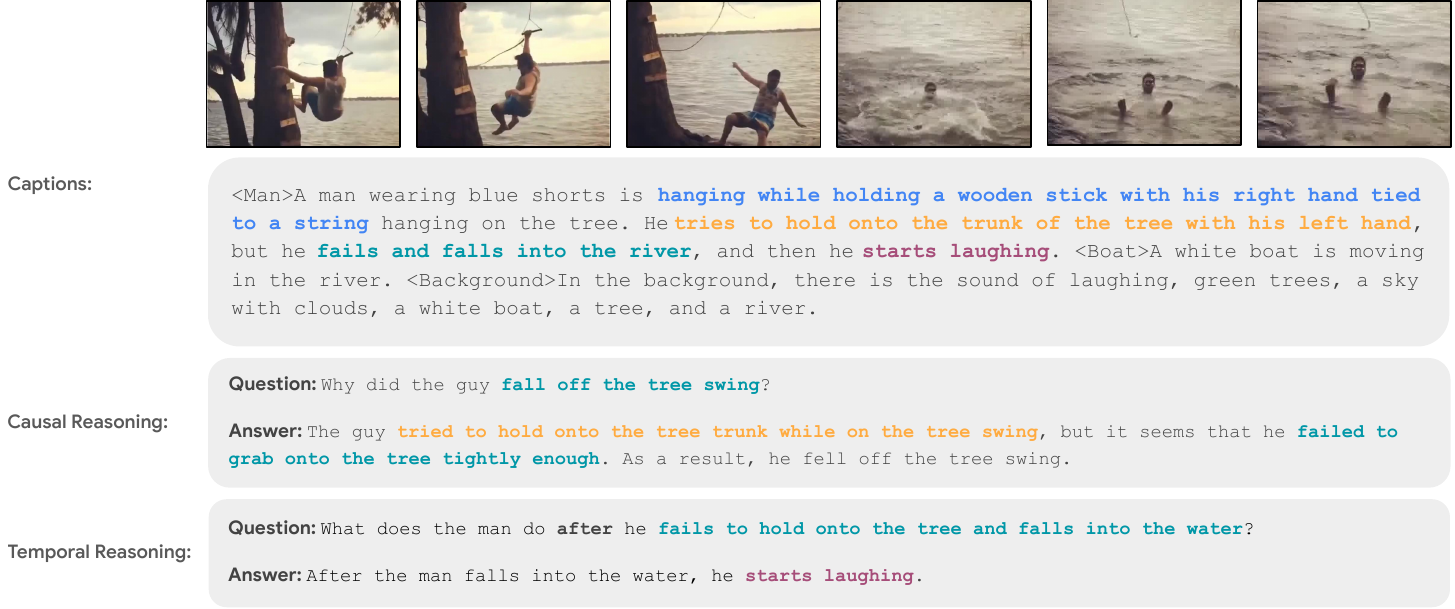}
\vspace{-5pt}
\caption{
\textbf{An example of the instruction-following data.} The first block shows the detailed captions used to prompt an LLM (PaLM~2~\cite{palm2} in our case), and the following two blocks show the LLM's responses. We show the keyframes in the top block for illustration purpose and do \textit{not} use them while prompting the LLM.
Different details in text are highlighted.
Best viewed in color.
}
\label{fig:instruction_examples}
\end{figure*}

\begin{figure*}[!tb]

\centering
\includegraphics[width=0.88\linewidth]{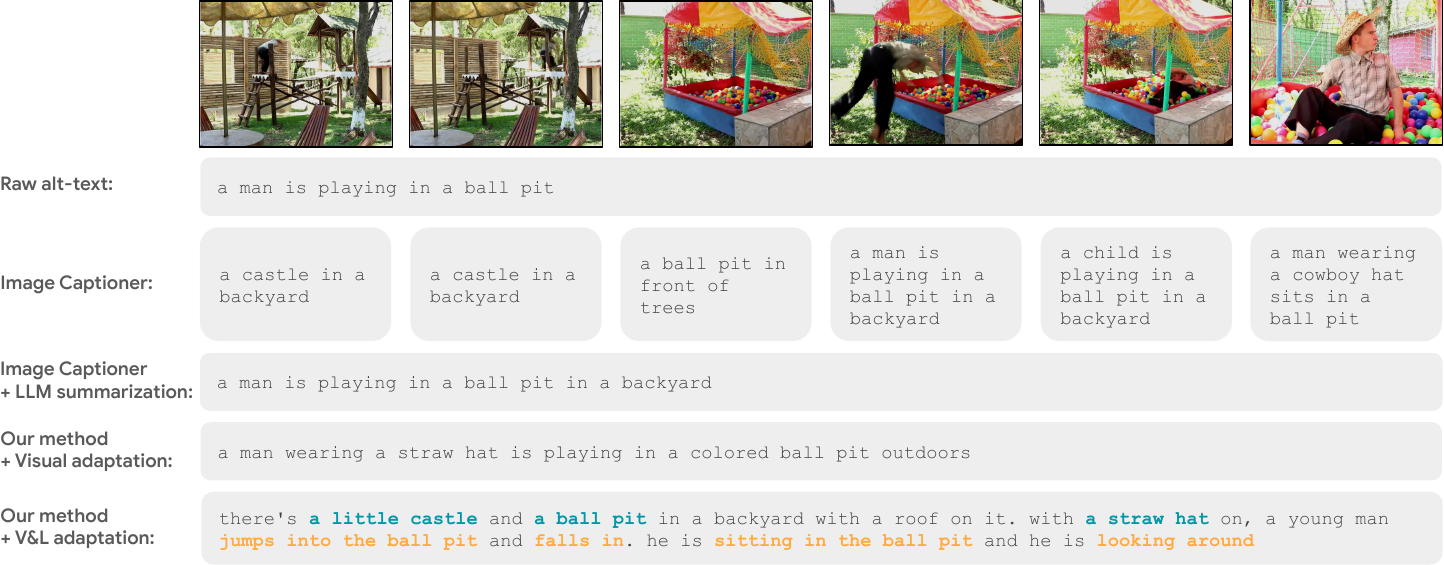}
\vspace{-5pt}
\caption{
\textbf{An example of video captions by PaLI-3 before and after video-specific adaptation.}
We show the keyframes on top for illustration purposes and the generated captions in the following blocks.
Different details in text are highlighted.
Best viewed in color.
}
\label{fig:caption_examples}
\vspace{-5pt}
\end{figure*}

\myparagraph{Inference.}
At inference time, we query the video-language model by feeding sampled video frames for $\vx$, the regular task prompt for captioning for $\vy$, and a special start-of-sentence token \texttt{<s>} for $\vz=[z_0]$.
We sample from the distribution recursively, i.e. $\tilde{z}_\ell \sim p(z|\vx,\vy,\tilde{z}_{<\ell})$ until an end-of-sentence token \texttt{</s>} is reached.
We use nucleus sampling~\cite{holtzman2020nucleus}, where we only sample from a subset of tokens that contain the vast majority of the probability mass at each step, multiple times.
We provide an example of captions before and after video-specific adaptation in~\Cref{fig:caption_examples}.
Readers can find more results in the supplementary materials in~\shortrefsec{supp:example}.
We observe on average 20\% longer length in the output sequence after the language adaptation while using the same task prompt for captioning.
We attribute it to the effectiveness of instruction tuning.

\section{Experiments}
\label{sec:exp}

First, we summarize the datasets that we use in~\shortrefsec{exp:datasets}. 
Next, we describe how we harness and evaluate the distilled pseudo-captions in~\shortrefsec{exp:harness}.
We show the main results,~\ie (1) the scaling effect of our data generation pipeline, (2) the quality of pseudo-captions by pre-training a dual-encoder model, and (3) the performance of the adapted video-language model on video-language tasks in~\shortrefsec{exp:main}.
Finally, we discuss the effect of different components in~\shortrefsec{exp:ablation}.

\subsection{Datasets}
\label{sec:exp:datasets}
\Cref{tab:dataset_summary} summarizes the video datasets used in this paper, and more details are in~\shortrefsec{supp:dataset} in the supplementary material. We categorize the datasets into four parts and describe the adaptation data and distilled data first.

\myparagraph{Adaptation data.}
We use two datasets to adapt a vision-language model from images to videos:
(1) \textit{Spoken Moments in Times (S-MiT)}~\cite{monfort2021smit} contains 500K videos with spoken captions.
The videos are typically short ($2\unsim3$ seconds) and the transcribed captions are brief (18 words on average per video).
It has 481K/8K/3K videos for training/validation/testing.
We use the training split to conduct visual adaptation of the video-language model and evaluate the video captioning result by CIDEr score on the testing split following PaLI~\cite{chen2023palix,chen2023pali3}.
(2) \textit{Video Localized Narratives (VidLN)}~\cite{voigtlaender2023vidln} annotates comprehensive events in videos which involve multiple actors and possibly actor-actor and actor-object interaction.
The narratives are longer (85 words on average) and are better suited to generate a diverse instructing-following corpus.
We use the training split which has 47,776 videos from the union of OVIS~\cite{qi2022ovis}, Oops~\cite{epstein2020oops}, UVO~\cite{wang2021uvo}, and Kinetics~\cite{carreira2017kinetics} datasets, to generate instruction-answer pairs for language adaptation.

\begin{table}
    \centering
    \tablestyle{3pt}{1.05}
    \resizebox{0.95\linewidth}{!}{
    \begin{tabular}{l|l|l|lH}
    Dataset & Task & Size & Metrics & Eval. Prot. \\
    \thickhline
    S-MiT~\cite{monfort2021smit} & ADP & 480K (train) & - & - \\
    VidLN~\cite{voigtlaender2023vidln} & ADP & 47K (train) & - & - \\
    \hline
    VideoCC~\cite{nagrani2022videocc} & CP & 7M{\color{gray}/10M} & - & - \\
    InternVid~\cite{wang2023internvid} & CP & 10M & - & - \\
    \hline
    MSR-VTT~\cite{xu2016msrvtt} & TVR & 1K (val, or \textit{1k-A}) & Recall@$k$ & ZS \\
    VATEX~\cite{wang2019vatex} & TVR & 1.5K (test as in~\cite{wang2022internvideo}) & Recall@1 & ZS \\
    Kinetics-600~\cite{carreira2018kinetics600} & CLS & 28K (val) & Accuracy & ZS \\
    \hline
    MSR-VTT~\cite{xu2016msrvtt} & CAP & 6.5K{\scriptsize(train)}+3K{\scriptsize(test)} & CIDEr & ZS, FT \\
    {\scriptsize MSR-VTT QA~\cite{xu2017msrvttqa}} & QA & 6.5K{\scriptsize(train)}+3K{\scriptsize(test)} & Accuracy & ZS, FT \\
    {\scriptsize ANet-Captions~\cite{krishna2017dense}} & CAP & 31K{\scriptsize(train)}+14K{\scriptsize(test)} & CIDEr & ZS, FT \\
    S-MiT~\cite{monfort2021smit} & CAP & 480K{\scriptsize(train)}+3K{\scriptsize(test)} & CIDEr & ZS, FT \\
    ANet-QA~\cite{yu2019activitynetqa} & QA & 32K{\scriptsize(train)}+8K{\scriptsize(test)} & Accuracy & ZS, FT \\
    \scriptsize NExT-OE-QA~\cite{xiao2021nextqa} & QA & 37K{\scriptsize(train)}+9K{\scriptsize(test)} & \scriptsize Wu-Palmer Similarity (WUPS) & {ZS, FT} \\
    \hline
    \end{tabular}
    }
    \vspace{-5pt}
    \caption{
    \small{
    \textbf{Dataset summary.}
    ADP is short for adapting VLMs while CP is for contrastive pre-training a dual-encoder model.
    Evaluation tasks include text-to-video retrieval (TVR), action classification (CLS), video captioning (CAP), and video question answering (QA).
    }}
    \label{tab:dataset_summary}
    \vspace{-10pt}
\end{table}

\myparagraph{Data with distilled pseudo-captions.}
We apply the resultant video-language model to caption two largest-scale webly-scraped video datasets:
(1) \textit{VideoCC}~\cite{nagrani2022videocc} contains $\unsim$10M video-caption pairs from 6M unique videos.
The raw alt-text is automatically retrieved from those in the Conceptual Captions image-captioning dataset (CC3M)~\cite{sharma2018conceptual} based on image similarity.
$\unsim$7.1M clips are available by the time of our experiments.
(2) \textit{InternVid}~\cite{wang2023internvid} has $\unsim$234M clips from 7M videos.
The original captions are synthesized from individual frames' captions by an LLM.
We use the publicly available InternVid-10M-FLT subset
which has 10M clips with top-scoring video-text similarities.
We denote the datasets processed by our method to be \textbf{VideoCC$^{+}$} and \textbf{InternVid$^{+}$}.
We use both datasets to pre-train a dual-encoder model to show the usefulness of the machine-generated video captions, explained next.

\subsection{Harnessing the Distilled Pseudo-Captions}
\label{sec:exp:harness}
We harness and \textit{evaluate} the distilled pseudo-captions for million-scale web-scraped videos, \textbf{VideoCC$^{+}$} and \textbf{InternVid$^{+}$}, using a dual-encoder model~\cite{radford2021clip}. The model's video understanding performance is a solid indicator of the pseudo-captions' quality, and we show that they are of higher quality than the original text in VideoCC and InternVid. 

\myparagraph{Contrastive training of a dual-encoder model.}
We train a video-language dual-encoder model like CLIP~\cite{radford2021clip}.
Specifically, given the input video frames $\vx$ and machine-generated captions $\tilde{\vc}$, the model applies a visual encoder $G_V$ plus a projection head $h_V$ and a text encoder $G_T$ plus a projection head $h_T$ in parallel to obtain the global visual and textual embedding, respectively,
{\small
\begin{align}
    \vu = h_V(G_V(\vx)), \vv = h_T(G_T(\tilde{\vc})).
\end{align}
}
We use the InfoNCE~\cite{oord2018infonce} loss to train the model. 
Note that we deliberately choose a different notation $G_{(\cdot)}$ than $F_{(\cdot)}$ in the VLM in~\shortrefsec{prelim} because the dual-encoder model does \emph{not} share any parameters with the VLM.

\myparagraph{Model architecture.}
The dual-encoder model has a vision encoder and a text encoder.
The video input is represented by 4 frames at 2 FPS.
The vision encoder is a Vision Transformer~\cite{dosovitskiy2021vit} with joint spatial-temporal attention (denoted as ``ViT-\textit{st}'') following~\cite{zhao2023training}.
We use ViT-L/14 to report the main result and ViT-B/16 for ablation studies if not otherwise specified.
The weights are initialized from CLIP~\cite{radford2021clip} except that we randomly initialize the temporal position embedding $\mathrm{PE}_t\in\mathbb{R}^{T\times D}$ and add it to the original spatial position embedding $\mathrm{PE}_s\in\mathbb{R}^{N\times D}$,~\ie
$\mathrm{PE}[i,:,:] = \mathrm{PE}_{t}[i,\texttt{None},:]+\mathrm{PE}_{s}[\texttt{None},:,:]$.
The text encoder is a 12-layer GPT-like Transformer~\cite{radford2019gpt2}.
It takes as input one video caption, tokenizes it using BPE~\cite{sennrich2016bpe}, and keeps at most 77 tokens.
If a video has more than one caption, we randomly sample one of them at each time.

\subsection{Main Results}
\label{sec:exp:main}

We report the dual-encoder model's text-to-video retrieval performance (on MSR-VTT and VATEX) and video classification accuracy (on Kinetics-600), both under the \textit{zero-shot} setting. These results are meant to evaluate the quality of the distilled video pseudo-caption data. Besides, we also evaluate the VLM adapted to the video domain on a few representative video-language benchmarks following PaLI-3~\cite{chen2023pali3}, including video captioning (MSR-VTT~\cite{xu2016msrvtt}, ActivityNet-Captions~\cite{krishna2017dense}) and video question-answering (MSR-VTT QA~\cite{xu2017msrvttqa}, ActivityNet QA~\cite{yu2019activitynetqa}, and NExT Open-Ended QA~\cite{xiao2021nextqa}).
We enumerate the datasets involved at the bottom of~\Cref{tab:dataset_summary} and leave details in~\shortrefsec{supp:dataset}.

\begin{figure}
\begin{center}
\begin{tikzpicture}
    \begin{axis} [
        axis x line*=bottom,
        axis y line*=left,
        legend pos=south east,
        legend columns=1,
        ymin=30, ymax=43,
        xmin=7, xmax=7000,
        width=0.7\linewidth,
        height=0.5\linewidth,
        ylabel={MSR-VTT R@1},
        xlabel={Number of used data in VideoCC ($\times 10^3$)},
        x tick label style = {anchor=north},
        symbolic x coords={7, 70, 700, 7000},
        xticklabel={\pgfmathparse{\tick}\pgfmathprintnumber{\pgfmathresult}},
        xtick=data,
        ymajorgrids = true,
        ytick={35, 40},
        ylabel style={align=center, font=\small, at={(-0.1, 0.5)}},
        xlabel style={font=\small},
        tick label style={font=\small},
        legend style={font=\footnotesize, at={(1.4,0.5)}, anchor=north},
        legend cell align={left},
    ]

    \addplot[mark=o,darkgreen!80!white,style={ultra thick}] plot coordinates {
        (7, 30.3)
        (70, 33.4)
        (700, 33.5)
        (7000, 33.5)
    };
    \addplot[mark=square,orange!80!white,style={ultra thick}] plot coordinates {
        (7, 30.6)
        (70, 33.1)
        (700, 34.0)
        (7000, 34.8)
    };
    \addplot[mark=triangle,blue,style={ultra thick}] plot coordinates {
        (7, 35.2)
        (70, 38.6)
        (700, 41.3)
        (7000, 42.1)
    };
    \addlegendentry{Alt-text~\cite{nagrani2022videocc}}
    \addlegendentry{Img. cap. + LLM}
    \addlegendentry{Vid. cap. (Ours)}
\end{axis}
\end{tikzpicture}
\end{center}
\vspace{-20pt}
\caption{
\textbf{Scaling effect of video captioning.}
For VLM-generated captions, the zero-shot video retrieval performance consistently improves with respect to an increasing amount of video data. 
Pre-training on retrieved alt-text quickly stagnates.
}
\label{fig:scaling}
\vspace{-10pt}
\end{figure}

\begin{table*}[!tb]
    \centering
    \tablestyle{4pt}{1.05}
    \begin{tabular}{l|l|ccc|ccc|HHcc}
    \multirow{2}{*}{Method} & \multirow{2}{*}{Pre-training Dataset}  & \multicolumn{3}{c|}{MSR-VTT TVR} & \multicolumn{3}{c|}{VATEX TVR} & \multicolumn{2}{H}{Kinetics-400} & \multicolumn{2}{c}{Kinetics-600} \\
    & & R@1 & R@5 & R@10 & R@1 & R@5 & R@10 & Top-1 & Top-5 & Top-1 & Top-5 \\
    \thickhline 
    CLIP~\cite{radford2021clip}            & WIT                 & 31.2 & 53.7 & 64.2 & 45.2 & - & - & 58.4 & 81.8 & 55.1 & 79.2 \\
    CLIP4Clip~\cite{luo2022clip4clip}      & WIT                 & 30.6 & 54.4 & 64.3 & - & - & - & - & - & - & - \\
    CLIP4Clip~\cite{luo2022clip4clip}      & WIT\to VideoCC (10M)   & 33.7 & 57.9 & 67.9 & - & - & - & - & - & - & - \\
    InternVideo~\cite{wang2022internvideo} & WIT\to Mixed (12M)     & 40.0 & 65.3 & 74.1 & 49.5 & 79.7 & 87.0 & - & - & - \\
    ViCLIP~\cite{wang2023internvid}        & WIT\to WebVid (10M)    & 35.6 & - & - & - & - & - & 59.9 & 82.1 & 58.7 & 81.0 \\
    ViCLIP~\cite{wang2023internvid}        & WIT\to InternVid (10M) & 42.4 & - & - & - & - & - & 64.8 & 86.6 & 62.2 & 84.9 \\
    \hline
    \multirow{7}{*}{CLIP (ViT-\textit{st}-L)} %
     & WIT\to VideoCC & 37.0 & 62.1 & 72.5 & 37.7 & 66.9 & 77.2 & 52.5 & 78.2 & 48.6 & 74.8 \\
    & WIT\to VideoCC$^{+}$ (\textbf{Ours}) & 48.2 & 72.2 & 80.8 & 64.2 & 90.2 & 95.1 & 64.0 & 87.8 & 61.1 & 85.6 \\
    & {Absolute gain $\Delta$} & \gain{+11.2} & \gain{+10.1} & \gain{+8.3} & \gain{+26.5} & \gain{+23.3} & \gain{+17.9} & \gain{+11.5} & \gain{+9.6} & \gain{+12.5} & \gain{+10.8} \\
    \cline{2-12}
    & WIT\to InternVid & 42.5 & 67.0 & 76.8 & 58.7 & 87.0 & 93.0 & 63.6 & 87.4 & 60.7 & 85.0 \\
    & WIT\to InternVid$^{+}$ (\textbf{Ours}) & 46.3 & 71.5 & 80.3 & 65.2 & 91.3 & 95.5 & 64.6 & {\textbf{88.2}} & 62.7 & 86.2 \\
    & {Absolute gain} $\Delta$ & \gain{+3.8} & \gain{+4.5} & \gain{+3.5} & \gain{+6.5} & \gain{+4.3} & \gain{+2.5} & \gain{+1.0} & \gain{+0.8} & \gain{+2.0} & \gain{+1.2} \\
    \cline{2-12}
    &  {WIT\to VideoCC$^{+}$+InternVid$^{+}$ (\textbf{Ours})} & {\textbf{48.4}} & {\textbf{73.5}} & {\textbf{81.9}} & {\textbf{65.6}} & {\textbf{91.7}} & {\textbf{95.8}} & {\textbf{65.4}} & {88.1} & {\textbf{62.8}} & {\textbf{86.4}} \\
    \hline
    \end{tabular}
    \vspace{-5pt}
    \caption{
    \small{
    \textbf{Zero-shot text-to-video retrieval performance on MSR-VTT \& VATEX and video recognition performance on Kinetics-600 using different sources of textual descriptions.}
    $\mathcal{D}^{+}$ means that the captions in the video dataset $\mathcal{D}$ are generated by our proposed pipeline.
    $\mathcal{D}\in\{\text{VideoCC}, \text{InternVid}\}$ in our experiments.
    }}
    \label{tab:main:dual_encoder}
    \vspace{-5pt}
\end{table*}

\begin{table*}
    \vspace{-5pt}
    \centering
    \tablestyle{5pt}{1.05}
    \begin{tabular}{l|l|cc|ccH|cH}
    \multirow{2}{*}{Method} & \multirow{2}{*}{Pre-training Dataset} & \multicolumn{2}{c|}{MSR-VTT} & \multicolumn{2}{c}{ActivityNet} & S-MiT  & NExT-OE-QA & Oops-QA \\
    & & Caption & QA (Acc.)  & Caption & QA (Acc.) & Caption  & QA (WUPS) & QA (Acc.) \\
    \thickhline
    \multirow{2}{*}{Prior SOTA} & \multirow{2}{*}{-} & 18.6 & 16.8 & 15.0 & 25.9 & - & 26.7 & 25.1 \\
    &  & {\scriptsize DeCap~\cite{li2023decap}} & {\scriptsize FrozenBiLM~\cite{yang2022frozenbilm}} & {\scriptsize  DeCap~\cite{li2023decap}} & {\scriptsize FrozenBiLM~\cite{yang2022frozenbilm}} & - & {\scriptsize Flamingo~\cite{alayrac2022flamingo}} & {\scriptsize PaLI~\cite{chen2023pali,voigtlaender2023vidln}} \\
    PaLI-3$_{8f}$~\cite{chen2023pali} & WebLI                        & 21.3 & 12.7 & 13.8 & 22.9  & 11.5 & 23.2 & 24.7 \\
    Ours                       & WebLI$\rightarrow$SMiT+VidLN & \textbf{48.2} & \textbf{24.4} & \textbf{31.0} & \textbf{29.6} & N/A & \textbf{29.5} & N/A \\
    \hline
    \end{tabular}
    \vspace{-5pt}
    \caption{
    \small{
    \textbf{Zero-shot performance of the Video-Language Model on video-language understanding tasks.}
    Our adapted video-language model significantly improves over the 8-frame PaLI-3 baseline and outperforms the best reported numbers. 
    }}
    \label{tab:main_vlm}
    \vspace{-5pt}
\end{table*}

\myparagraph{Distilled vs.\ alt-text captions at various scales.}
\label{sec:exp:scaling}
\Cref{fig:scaling} shows that the distilled pseudo-captions for VideoCC outperform VideoCC's original Alt-text captions, by a striking margin, when the dual-encoder models trained using different subsets of VideoCC are evaluated on the MSR-VTT retrieval task. %
We find that Recall@1 quickly saturates when training the dual-encoder model on VideoCC with alt-text.
Specifically, training with only 1\% VideoCC$^+$ ($\unsim$70K) achieves the same level of Recall@1 with training with the whole VideoCC set ($\unsim$7M), indicating that the original alt-text scales poorly.
We attribute the alt-text's inferior performance to a compounding error of textual noise~\cite{jia2021align}, spurious correlation when computing visual similarities~\cite{yang2023mitigating}, and the visual discrepancy between images and videos.
In contrast, training the dual-encoder model with the pseudo-captions clearly exhibits a pleasant scaling effect: R@1 consistently increases with more pre-training video data.
We also include in~\Cref{fig:scaling} the curve corresponding to the pseudo-captions distilled from the image-language model before it is adapted to the video domain. It almost overlaps with the alt-text curve at the beginning and then becomes slightly better near the end.

\myparagraph{Distilled captions for video understanding.}
We continue to evaluate the distilled pseudo-captions by the corresponding dual-encoder model's \textit{zero-shot} performance on text-to-video retrieval and video classification.
From~\Cref{tab:main:dual_encoder}, we see that the pseudo-captions distilled from our VLM significantly improve the dual-encoder over the original text in VideoCC and InternVid.
On VideoCC, with all other settings being the same, the dual-encoder model trained on VideoCC$^{+}$, achieves 48.2\% Recall@1 on MSR-VTT, 11.2\% better than the one trained on the original Alt-text.
It also clearly surpasses the recent ViCLIP trained on InternVid, which contains $2\times$ more unique videos than VideoCC.
On InternVid, our model trained on InternVid$^{+}$ is 3.8\% better than the baseline trained on the original InternVid's auto-generated captions.
It is worth noting that our adapted VLM is also ``lighter-weight'' compared to the multi-scale captioning pipeline in InternVid~\cite{wang2023internvid}, which relies on both image captioning models (BLIP-2)~\cite{li2023blip2} on multiple frames and an LLM to put them together.
We also highlight the zero-shot top-1 and top-5 classification accuracy on Kinetics-600.
For instance, the dual-encoder model trained on VideoCC$^{+}$/InternVid$^{+}$ improves the baselines on VideoCC/InternVid by 12.5/2.0\% top-1 accuracy.

Interestingly, we notice that the model trained on InternVid$^{+}$ works better on action recognition, while the one trained on VideoCC$^{+}$ is better on video retrieval.
This is probably because the InternVid videos are specifically collected based on action phrases~\cite{wang2023internvid}, while VideoCC is seeded from image-captioning data~\cite{nagrani2022videocc}.
Since the two datasets are complementary, combining them indeed leads to performance gains as shown in the last row in \Cref{tab:main:dual_encoder}.

\myparagraph{Evaluating the video-language model.}
We compare the adapted VLM with the baseline PaLI-3 in~\Cref{tab:main_vlm}.
We focus on the zero-shot performance where we apply the model to the testing split of downstream tasks \textit{without} any tuning.
This setting resembles the scenario where we generate pseudo-captions on VideoCC and InternVid, and it provides us with a direct measure on well-established benchmarks. Specifically, the greatly improved CIDEr score on MSR-VTT and ActivityNet-Captions showcases the effectiveness of adapting a VLM to the video domain.
We also see excellent zero-shot question-answering results compared to PaLI-3.
On the challenging open-ended NExT-QA dataset, our model outperforms Flamingo~\cite{alayrac2022flamingo} by 2.8\% (WUPS score).
This gain is achieved using only $\frac{1}{16}\times$ of the parameters (5B \vs 80B) and $\frac{1}{50}\times$ of training videos (0.55M publicly available S-MiT\&VidLN \vs 27M in-house VTP).
On MSR-VTT QA and ActivityNet QA, our adapted model achieves 7.6\% and 3.7\% higher accuracy than FrozenBiLM~\cite{yang2022frozenbilm}, trained on WebVid-10M~\cite{bain2021webvid}.

\subsection{Ablation Studies}
\label{sec:exp:ablation}

\vspace{-2pt}
\myparagraph{What makes captioning better?}
We investigate the key to generating better captions for contrastive pre-training video-language dual-encoder models in~\Cref{tab:ablate:captions}.
The comparison starts from the alt-text-only baseline which achieves 37.0\% text-to-video R@1 on MSR-VTT.
Using frame-level captions produced by PaLI-3 \textit{as-is} increases R@1 by 2.5\%.
We also attempt to merge multiple frames' captions into a single sentence with PaLM-2~\cite{palm2} similar to the pipeline in InternVid~\cite{wang2023internvid} but see marginal gain (0.3\%).
This result is consistent with our observation that LLMs often fail to interpolate when key temporal information is lost in the image-level descriptions.
We also encounter a trade-off between being concise but lacking diversity and being detailed but vulnerable to hallucination.
If we conduct visual adaptation in PaLI-3, the resulting video captions almost double the gain from 2.5\% to 4.7\%.
Generating multiple captions independently with nucleus sampling contributes 1.9\%.
Finally, doing language adaptation on PaLI-3 with instruction-following data further improves R@1 by 0.7\%. 

\begin{table}[!tb]
    \centering
    \tablestyle{2pt}{1.05}
    \resizebox{0.85\linewidth}{!}{
    \begin{tabular}{c|c|c|c|c||l}
        \multirow{2}{*}{PaLI-3} & \multirow{2}{*}{LLM} & \multicolumn{2}{c|}{Adapting VLM~(\shortrefsec{method:adaptation})} & {Multi.} & \multicolumn{1}{c}{MSR-VTT} \\
        & & Visual & Language & {Samples} & \multicolumn{1}{c}{Recall@1} \\
        \thickhline
        & & & & & 37.0 \\
        \cmark & & & & & 39.5\gain{(+2.5)} \\
        \cmark & \cmark & & & & 39.8\gain{(+2.8)} \\
        \cmark & & \cmark & & & 41.7\gain{(+4.7)} \\
        \cmark & & \cmark & & \cmark & 43.6\gain{(+6.6)} \\
        \cmark & & \cmark & \cmark & \cmark & 44.3\gain{(+7.3)} \\
        \hline
    \end{tabular}
    }
    \vspace{-5pt}
    \caption{
    \textbf{The effect of using different sources of textual descriptions.}
    The captioning quality is measured by the zero-shot text-to-video retrieval performance (Recall@1) on MSR-VTT.
    The first line with no components checked refers to the alt-text baseline.
    The ``LLM''-column means that we use PaLM 2~\cite{palm2} to summarize captions from multiple frames similar to~\cite{wang2023internvid}.
    }
    \vspace{-5pt}
    \label{tab:ablate:captions}
\end{table}

\begin{table}[!tb]
    \centering
    \vspace{-5pt}
    \tablestyle{3pt}{1.05}
    \begin{tabular}{c|c|c|c}
    \multicolumn{3}{c|}{Visual Adaptation} & S-MiT Caption \\
    $F_V$ & Self-training & $F_L$ & (CIDEr) \\
    \thickhline
    \xmark & & \cmark & 41.2 \\
    \cmark & & \xmark & 42.3 \\
    \cmark & & \cmark & 40.3 \\
    \cmark & \cmark & \xmark & 43.5 \\
    \hline
    \end{tabular}
    \vspace{-5pt}
    \caption{
    \small{
    \textbf{Adapting vision encoder.} $\cmark$ and $\xmark$ denote fine-tuning and freezing the parameters respectively.
    Fine-tuning the visual part while freezing the language model yields better results.
    }}
    \label{tab:ablate:visual_adaptation}
    \vspace{-10pt}
\end{table}

\myparagraph{How should we do visual adaptation?}
We study several ways for visual adaptation in~\Cref{tab:ablate:visual_adaptation}.
The first option,~\ie freezing the visual encoder $F_V$ while fine-tuning the language model $F_L$, takes inspiration from LiT~\cite{zhai2022lit}.
This leads to a drop of 1.1 CIDEr score compared to our default recipe, where $F_V$ is fine-tuned and $F_L$ frozen.
We ascribe it to the visual discrepancy between images and videos: The downstream tasks in~\cite{zhai2022lit,chen2023pali} are mostly still images, the same as the large-scale pre-training data.
In contrast, the videos of our interests have unique characteristics such as object movement, camera motion, and the resultant visual degradation.
We also observe a performance drop if we fine-tune both $F_V$ and $F_L$.
This recipe may be prone to over-fitting because the video-text dataset lacks diversity and quantity.
Finally, we show that self-training with VideoCC pseudo-captions (details in~\shortrefsec{supp:self_train}) improves captioning results by 1.2 CIDEr score, reaching 43.5.
It is worth noting that this number is on par with the best-performing PaLI-X~\cite{chen2023palix} which has $11\times$ more parameters and takes $2\times$ more frames as input than ours.

\begin{table}[!tb]
    \centering
    \tablestyle{3pt}{1.05}
    \resizebox{0.96\linewidth}{!}{
    \begin{tabular}{l|c|c|c}
    \multirow{2}{*}{Instruction data} & MSR-VTT & ActivityNet & NExT-OE \\
    & Caption {\scriptsize (CIDEr)} & Caption {\scriptsize (CIDEr)} & QA {\scriptsize (WUSP)} \\
    \thickhline
    None (PaLI-3) & 21.3 & 13.8 & 23.2 \\
    LLaVA 1.0~\cite{liu2023llava} & 16.9 & 25.1 & 16.3 \\
    ActivityNet-Instruct~\cite{maaz2023videochatgpt} & 30.8 & 34.6 & 11.7 \\
    \hline
    Ours & & & \\
    {\scriptsize \quad + VidLN Causal/temporal Reasoning} & 28.5 & 29.5 & 5.0 \\
    {\scriptsize \quad + SMiT Captions} & \textbf{51.6} & \textbf{35.1} & 3.9 \\
    {\scriptsize \quad + VidLN Short-QA} & 48.2 & 31.0 & \textbf{29.5} \\
    \hline
    \end{tabular}
    }
    \vspace{-5pt}
    \caption{
    \small{
    \textbf{Effect of instruction data.}
    Our proposed instruction data benefits the adaptation of the video-language model, reflected by better zero-shot captioning results and QA accuracy.
    }}
    \label{tab:ablate:instruct_data}
    \vspace{-10pt}
\end{table}

\myparagraph{How should we do language adaptation?}
We study the effect of instruction-following data in~\Cref{tab:ablate:instruct_data} when doing language adaptation.
We start with some representative visual instructional tuning datasets.
The first is LLaVA-1.0~\cite{liu2023llava} with 150K instruction pairs.
We find that it improves the CIDEr score by 7.3 on ActivityNet Captions but decreases by 4.4 on MSR-VTT Captions.
The second is ActivityNet-Instruct~\cite{maaz2023videochatgpt} with 100K instruction pairs from ActivityNet-Captions~\cite{krishna2017dense}.
It improves CIDEr score on both MSR-VTT and ActivityNet Captions, indicating that video-specific instructional-following data is essential to video-language tasks.
We then conduct an incremental study on our LLM-prompted instructional corpus on VidLN+SMiT by adding one component at a time.
First, we fine-tune the language model with only reasoning data.
The adapted model works on par with the one fine-tuned on ActivityNet-Instruct on ActivityNet-Captions even without seeing ActivityNet videos, demonstrating the generalization of our instructed data.
Next, we include the captioning data on S-MiT and see a higher CIDEr score on MSR-VTT and ActivityNet Caption.
However, both models suffer from significant degradation in zero-shot QA accuracy.
This is expected since the answers in all existing video QA datasets are typically short ($1\unsim3$ words) while our instructional data typically contains detailed reasoning (\Cref{fig:instruction_examples}).
To mitigate the gap, we further add QA pairs that are few-shot prompted based on Oops-QA~\cite{voigtlaender2023vidln}, and prepend the question with a QA-specific task prompt (``\texttt{Answer in en:}'').
The final model restores its zero-shot question-answering ability at the cost of a slight performance drop in captioning.

\myparagraph{More ablations.} We leave more ablations and discussions in the supplementary materials.
\section{Conclusion}
\label{sec:conclusion}

We present an approach to adapting an image-based vision-language model to videos and distilling high-quality pseudo-captions for millions of videos.
The adapted video-language model obtains excellent zero-shot performance on various video-language benchmarks.
The pseudo-captions yield a stronger dual-encoder model and show positive scaling behavior with respect to the number of videos.

\myparagraph{Acknowledgements.}
{This material is based upon work in part supported by the National Science Foundation under Grant No. IIS-1845485.}

{
    \small
    \bibliographystyle{ieeenat_fullname}
    \bibliography{main}
}

\clearpage
\appendix

\section{Instruction-Following Templates}
\label{sec:supp:template}
We provide the templates to generate the instruction-following data in~\Cref{table:supp:instruction_prompts}.
Specifically, each prompt starts with a brief instruction followed by two examples for few-shot prompting.
The examples used for prompting temporal-reasoning, causal-reasoning, and short question-answer pairs are enumerated in~\Cref{table:supp:instruction_temporal_examples},~\Cref{table:supp:instruction_causal_examples}, and~\Cref{table:supp:instruction_short_examples}, respectively.
We randomly sample two examples out of three at each time.

\section{Examples of Video Captioning}
\label{sec:supp:example}
\begin{figure*}[!tb]
\begin{subfigure}[b]{1.0\textwidth}
\centering
\includegraphics[width=0.9\linewidth]{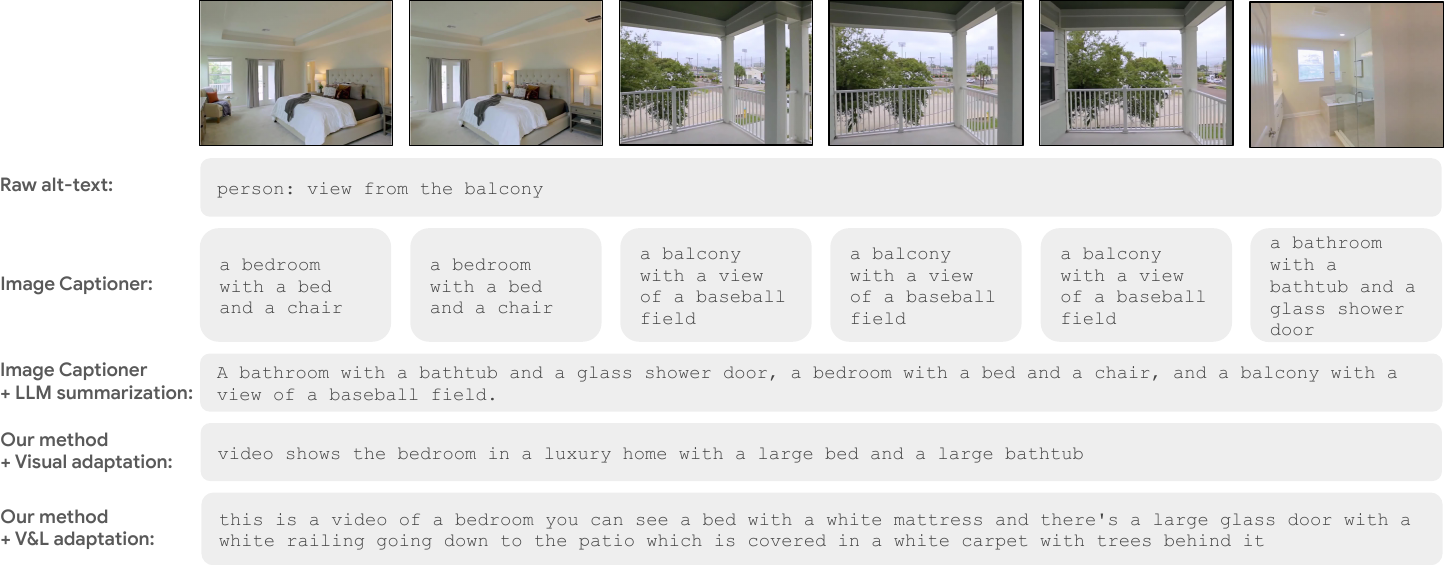}
\end{subfigure}
    
\noindent\rule{\linewidth}{0.4pt}

\begin{subfigure}[b]{1.0\textwidth}
\centering
\includegraphics[width=0.9\linewidth]{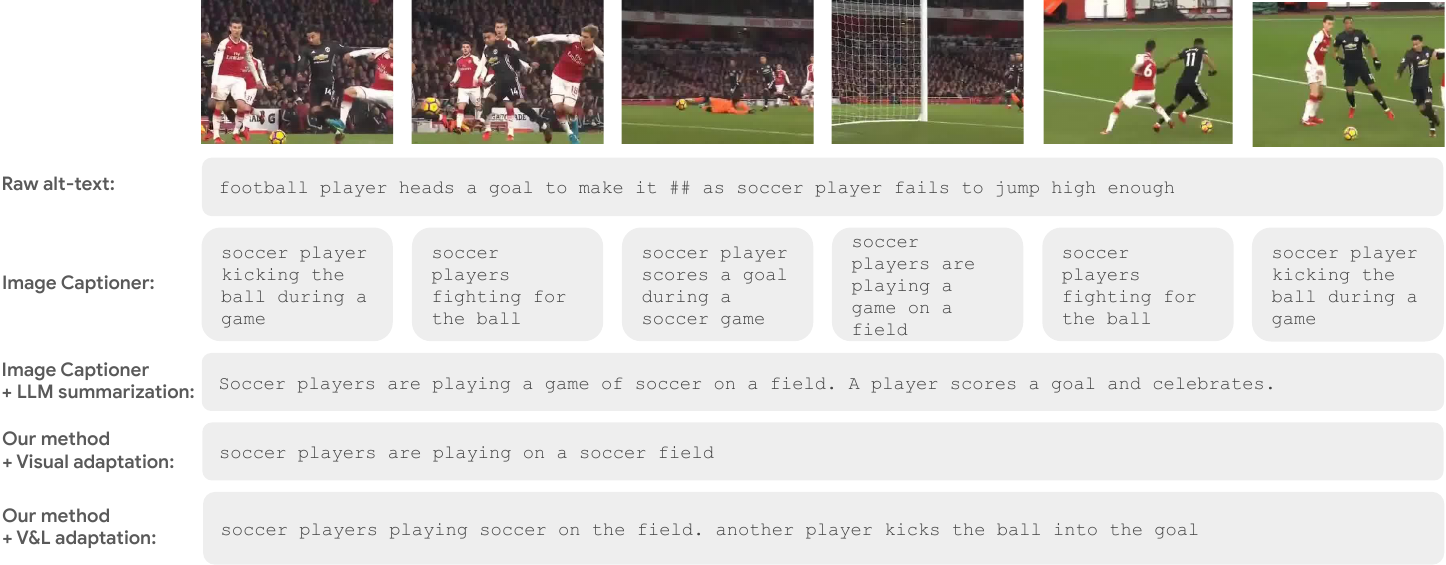}
\end{subfigure}

\noindent\rule{\linewidth}{0.4pt}

\begin{subfigure}[b]{1.0\textwidth}
\centering
\includegraphics[width=0.9\linewidth]{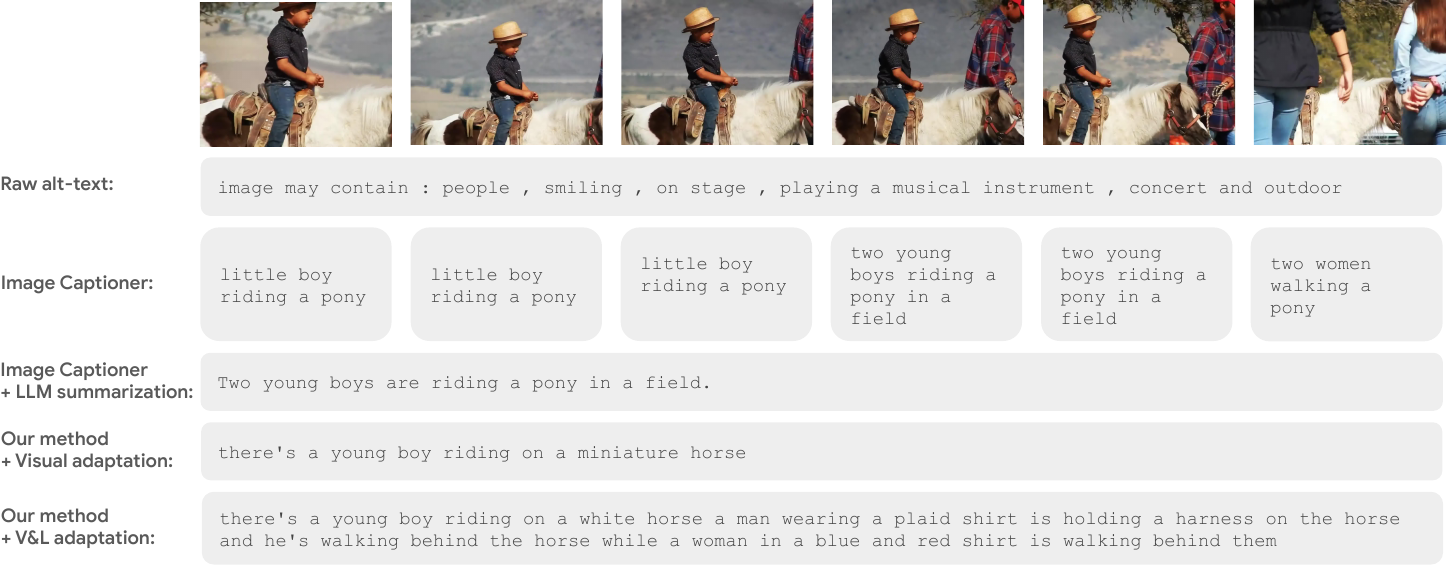}
\end{subfigure}
\vspace{-5pt}
\caption{
\textbf{More examples of video captions by PaLI-3 before and after video-specific adaptation.}
We show the keyframes on top for illustration purposes and the generated captions in the following blocks.
Different details in text are highlighted.
Best viewed in color.
}
\label{fig:supp:caption_examples}
\vspace{-5pt}
\end{figure*}

We provide some more examples of captions before and after video-specific adaptation in~\Cref{fig:supp:caption_examples}.
We can see that our video-language model with visual adaptation generates short and accurate captions for videos.
The outcome is comparable to the one that is achieved by frame-level image captioning following by LLM-summarization.
Furthermore, our video-language model with both visual and language adaptation provides more details when describing the same video.

Additionally, we show some representative statistics from the natural language understanding perspective in~\Cref{table:supp:stats}.
We analyze the sentence length and frequency of unique verbs/nouns using Spacy~\cite{spacy2020}.
Our generated video captions (VideoCC+) are nearly 50\% longer than alt-text.
They also contain more unique verbs and nouns than the original alt-text, indicating higher diversity.

\section{Dataset Details}
\label{sec:supp:dataset}

In this section, we summarize the datasets that we used in~\shortrefsec{exp} to evaluate the video-language model and the dual-encoder model.
The datasets that we used to adapt the vision-language model from images to videos and distill the resultant video-language model for pseudo-captioning have already been summarized in~\shortrefsec{exp:datasets}.

\subsection{Data for Evaluating the Dual-Encoder Model}

\myparagraph{MSR-VTT}~\cite{xu2016msrvtt} consists of 10K video clips with video captioning, each of which has 20 captions.
We follow the 1k-A splits in~\cite{wang2022internvideo}, namely 9K/1K for training/testing, and report text-to-video retrieval (TVR) Recall@\{1,5,10\} on the testing split.

\myparagraph{Kinetics-600}~\cite{carreira2018kinetics600} contains around 480K 10-second video clips from 600 action classes. 
We follow the standard splits, namely 390K/30K/ 60K for training/validation/testing, and report top-1 accuracy on the validation split.

\myparagraph{VATEX}~\cite{wang2019vatex} consists of around 41K videos sampled from the Kinetics-600 dataset, each of which has 10 English captions and 10 Chinese captions.
Only the English annotations are used for evaluation following~\cite{wang2022internvideo,yan2022videococa}.
We follow the splits in~\cite{wang2022internvideo}, namely 26K/1.5K/1.5K for training/validation/testing, and report text-to-video retrieval (TVR) Recall@\{1,5,10\} on the testing split.

\begin{table}[!tb]
\tablestyle{2pt}{1.05}
\begin{tabular}{c|c|c|c}
& Average length & \# of uniq. verbs & \# of uniq. nouns \\
\thickhline
VideoCC & 10.66 & 8,000 & 13,097 \\
VideoCC$^{+}$ & 15.74 & 8,317 & 29,712 \\
\hline
\end{tabular}
\caption{
\textbf{Statistics of original alt-text and pseudo-captions.}}
\label{table:supp:stats}
\end{table}

\subsection{Data for Evaluating the Video-Language Model}

\myparagraph{MSR-VTT Captions}~\cite{xu2016msrvtt} consists of 10K video clips with video captioning, each of which has 20 captions.
We follow the standard splits in~\cite{xu2016msrvtt}, namely 6.5K/0.5K/3K for training/validation/testing, and report captioning results measured by CIDEr score on the testing split.

\myparagraph{ActivityNet Captions} consists of 100K temporally localized sentences for 20K
videos.
We follow the standard splits in~\cite{krishna2017dense}, namely 10K/5K/5K videos for training/validation/testing, and assume ground truth temporal proposals is known at evaluation.
We report captioning results measured by CIDEr score on \texttt{val\_2} split.

\myparagraph{MSR-VTT-QA}~\cite{xu2017msrvttqa} has the same amount of videos of MSR-VTT but is augmented with 243K question-answer pairs.
We follow the standard splits in~\cite{xu2017msrvttqa}, namely 158K/12K/73K QA pairs for training/validation/testing.
We report the accuracy (using exact string match as in PaLI~\cite{chen2023palix}) on the testing split.

\myparagraph{ActivityNet-QA}~\cite{yu2019activitynetqa} builds upon ActivityNet and contains 58K question-answer pairs.
We follow the standard splits, namely 32K/18K/8K QA pairs for training/validation/testing.
We report accuracy (using exact string match as in PaLI~\cite{chen2023palix,chen2023pali3}) on the testing split.

\myparagraph{NExT-OE-QA}~\cite{xiao2021nextqa} is the open-ended task for NExT-QA dataset.
It contains 52,044 question-answer pairs for a total of 5,440 videos.
Following~\cite{xiao2021nextqa}, we report Wu-Palmer Similarity (WUPS) score on the test set, which has 9.2K QA pairs.

\begin{table*}[!tb]
    \centering
    \tablestyle{4pt}{1.05}
    \begin{tabular}{l|l|ccc|ccc|HHcc}
    \multirow{2}{*}{Method} & \multirow{2}{*}{Pre-training Dataset}  & \multicolumn{3}{c|}{MSR-VTT TVR} & \multicolumn{3}{c|}{VATEX TVR} & \multicolumn{2}{H}{Kinetics-400} & \multicolumn{2}{c}{Kinetics-600} \\
    & & R@1 & R@5 & R@10 & R@1 & R@5 & R@10 & Top-1 & Top-5 & Top-1 & Top-5 \\
    \thickhline 
    InternVideo~\cite{wang2022internvideo} & WIT\to Mixed (12M)     & 40.0 & 65.3 & 74.1 & 49.5 & 79.7 & 87.0 & - & - & - \\
    ViCLIP~\cite{wang2023internvid}        & WIT\to WebVid (10M)    & 35.6 & - & - & - & - & - & 59.9 & 82.1 & 58.7 & 81.0 \\
    ViCLIP~\cite{wang2023internvid}        & WIT\to InternVid (10M) & 42.4 & - & - & - & - & - & 64.8 & 86.6 & 62.2 & 84.9 \\
    \hline
    \multirow{4}{*}{CLIP (ViT-\textit{st}-L)} & WIT\to S-MiT & {45.2} & {70.8} & {80.5} & \textbf{66.7} & \textbf{92.0} & \textbf{96.2} & \textbf{67.4} & \textbf{90.7} & \textbf{64.2} & \textbf{88.8} \\
    & WIT\to VideoCC$^{+}$ (\textbf{Ours}) & 48.2 & 72.2 & 80.8 & 64.2 & 90.2 & 95.1 & 64.0 & 87.8 & 61.1 & 85.6 \\
    & WIT\to InternVid$^{+}$ (\textbf{Ours}) & 46.3 & 71.5 & 80.3 & 65.2 & 91.3 & 95.5 & 64.6 & {\textbf{88.2}} & 62.7 & 86.2 \\
    &  {WIT\to VideoCC$^{+}$+InternVid$^{+}$ (\textbf{Ours})} & {\textbf{48.4}} & {\textbf{73.5}} & {\textbf{81.9}} & {\underline{65.6}} & {\underline{91.7}} & {\underline{95.8}} & {\underline{65.4}} & {\underline{88.1}} & {\underline{62.8}} & {\underline{86.4}} \\
    \hline
    \end{tabular}
    \caption{
    \small{
    \textbf{Comparison of zero-shot text-to-video retrieval performance on MSR-VTT \& VATEX and video recognition performance on Kinetics-600 between human-labeled and pseudo-captioned videos.}
    $\mathcal{D}^{+}$ means that the captions in the video dataset $\mathcal{D}$ are generated by our proposed pipeline.
    $\mathcal{D}\in\{\text{VideoCC}, \text{InternVid}\}$ in our experiments.
    }}
    \label{tab:supp:dual_encoder}
\end{table*}

\section{Implementation Details}
\label{sec:supp:implement}

\subsection{Adapting the Vision-Language Model}
We inherit the training recipe of PaLI-3 when adapting the vision-language model from images to videos.
Specifically, we use AdaFactor with $\beta_1=0$ and $\beta_2=0.8$. 
For the learning rate schedule, we use a linear warmup at the first 1K iteration, followed by inverse square-root decay.
The peak learning rate is $10^{-4}$.
During visual adaptation, we use a batch size of 64 and train the model for 40K iteration, which is equivalent to $\unsim$5 epochs, on 128 TPU-v5e chips.
During language adaption, we use a batch size of 256 and train the model for 10K iteration, which is equivalent to $\unsim$2.6 epochs, on 128 TPU-v5e chips.

\subsection{Training the Dual-Encoder Model}
We use SGD with momentum $\beta=0.9$ as the optimizer by default.
For the learning rate schedule, we use a linear warmup at the first 5K iterations followed by cosine decay.
The peak learning rate is $10^{-3}$.
We use a batch size of 1,024 and train the model for 100 epochs.
Besides, we observe that using the AdamW optimizer gives a faster convergence speed and a better final zero-shot performance when training the dual-encoder model with pseudo-captions.
Therefore, we use AdamW with $(\beta_1, \beta_2)=(0.9, 0.999)$ and weight decay of $0.01$ and train the model for 20 epochs when reporting the main result in~\Cref{tab:main:dual_encoder}.
We use the default SGD-optimizer recipe in~\Cref{tab:ablate:captions}.
For data augmentation, we apply standard scale jittering augmentation with a scale range of $(0.9, 1.33)$ and take a $224\times 224$ crop.

\section{More ablations}

\subsection{Justifying the Two-Stage Adaptation Design}
We choose the adaptation order based on the information flow of the system,~\ie the Vision-Language Model needs to first perceive well in order to speak correctly.
Thus we first adapt visual encoder to videos and then instruct-tune language model for details and completeness.
We conduct experiments of reversing the order in~\Cref{tab:supp:adaptation_order} and see a noticeable drop of WUPS (29.5$\rightarrow$17.9).

We also compare prompt-tuning (PT)~\cite{lester2021pet}, an adapter-based fine-tuning method during language adaptation.
Though the captioning result is close, the zero-shot QA accuracy is much worse.
This indicates that adapter methods are not aimed at general-purpose multi-modal models.

\begin{table}
\tablestyle{1.5pt}{1.05}
\begin{tabular}{c|c|c}
& S-MiT (CIDEr) & NExT-QA (WUPS) \\
\thickhline
V $\rightarrow$ L & 42.3 & 29.5 \\
L $\rightarrow$ V & 41.9 & 17.9 \\
V $\rightarrow$ L (PT) & 40.9 & 4.88 \\
\hline
\end{tabular}
\vspace{-5pt}
\caption{
\textbf{Effect of different adaptation orders.}}
\label{tab:supp:adaptation_order}
\end{table}

\subsection{Effect of Data Scale for Adaptation}
We first describe the criteria of choose training datasets in our adaptation.
The goal of visual adaptation is to align video features  with language, especially actions that hardly exist in images.
This calls for unambiguous and clean video-text pairs.
The goal of language adaptation is to learn a general-purpose video-language model from videos with detailed text annotations,~\ie narratives.
We thus choose the best available datasets to serve these two goals: S-MiT is the largest human-annotated video captioning dataset;
VidLN is the largest and most diverse video dataset with detailed human narratives.

We study the effect of the dataset size in~\Cref{tab:supp:smit_size} and \Cref{tab:supp:vidln_size}.
Following the motivations above, we measure the visual adaptation performance by supervised video captioning and the language adaptation performance by zero-shot QA.
For both stages, our model benefits from more data.
This justifies the scaling ability of our model.

\begin{table}
\tablestyle{3pt}{1.05}
\begin{tabular}{c|cccc}
\% of S-MiT videos & 0 & 1\% & 10\% & 100\% \\
\thickhline
S-MiT (CIDEr) & 11.5 & 37.6 & 39.6 & 42.3 \\
\hline
\end{tabular}
\vspace{-5pt}
\caption{
\textbf{Effect of data size in visual adaptation.}}
\label{tab:supp:smit_size}
\end{table}

\begin{table}
\tablestyle{3pt}{1.05}
\begin{tabular}{c|cccc}
\% of VidLN videos  & 0 & 1\% & 10\% & 100\% \\
\thickhline
NExT-QA (WUPS) & 1.2 & 8.8 & 16.0 & 29.5\\
\hline
\end{tabular}
\vspace{-5pt}
\caption{
\textbf{Effect of data size in language adaptation.}}
\label{tab:supp:vidln_size}
\end{table}

\subsection{Self-training with Pseudo-captioned Videos}
\label{sec:supp:self_train}
The generated captions along with the videos can be used to further improve the VLM via self-training.
We do this in the stage of visual adaptation because the language adaptation stage is mainly fueled by instruction-following data and adding pseudo-captioning leads to potential model drifting.
Let $\mathcal{D}_l=\{(\vx, \vc)\}$ and $\mathcal{D}_u=\{(\vx, \tilde\vc)\}$ denote the set of human-captioned videos and VLM-captioned videos respectively.
In each step, we construct a training batch by sampling a few samples from both sets, namely $\mathcal{B}=\mathcal{B}_u\cup\mathcal{B}_l$, where $\mathcal{B}_l\subset\mathcal{D}_l$ and $\mathcal{B}_u\subset\mathcal{D}_u$.
Compared to self-training with ``pseudo-labels'',~\ie either manually assigned one-hot targets after filtering~\cite{ghadiyaram2019large,duan2020omnisource} or output logits~\cite{berthelot2019mixmatch,sohn2020fixmatch}, pseudo-captioning provides richer supervision and naturally handles the long-tail issue.

\subsection{Comparing with human-labeled data}
In this section, we compare the performance of the dual-encoder model trained on the human-labeled data and pseudo-captions.
We train a dual-encoder model on the human-labeled S-MiT because (1) it is the largest human-labeled video-caption dataset to date and (2) our video-language model that is used to generate pseudo-captions for unlabeled videos is trained on S-MiT first.
The zero-shot retrieval and recognition performance is reported in~\Cref{tab:supp:dual_encoder} in comparison with the result on VideoCC$^{+}$ and InternVid$^{+}$.
We can see that the dual-encoder model trained on both VideoCC$^{+}$ and InternVid$^{+}$ clearly outperforms the one trained on S-MiT in terms of MSR-VTT zero-shot text-to-video retrieval recall.
This indicates that our adapted video-language model not only distills human-labeled video dataset, but also generalizes to unseen videos that are within the same domain.
When looking at the retrieval result on VATEX and classification result on Kinetics-600, the dual-encoder model trained on either VideoCC$^{+}$ or InternVid$^{+}$, however, is slightly inferior to that on S-MiT.
We ascribe this to the semantic correlation between S-MiT and Kinetics/VATEX: S-MiT is built on top of Moments-in-Times (MiT) whose videos are all tagged with one action or activity label similar to the way Kinetics and VATEX is constructed and the action concepts between MiT and Kinetics are closely related.

\begin{table*}[!tb]
\begin{mdframed}[align=center,linewidth=1pt]
{\textcolor{blue}{Temporal reasoning}}

You are an AI visual assistant that can analyze a single video. You receive a few sentences, each describing the same video you are observing.
The task is to use the provided caption, create a plausible question about the video, and provide the answer in detail.\newline
\textbf{Create questions that requires reasoning about temporal relationships between actions, determined by order of occurrence. The questions can also cover interactions between different persons or objects.}\newline
To answer such questions, one should require first understanding the visual content, then based on the background knowledge or reasoning, either explain why the things are happening that way, or provide guides and help to user's request.  Make the question challenging by not including the visual content details in the question so that the user needs to reason about that first.\newline
Always answer as if you are directly looking at the video.\newline
(1) Captions:\n\{\}\n\newline
Generated QA:\n\{\}\n\newline
(2) Captions:\n\{\}\n\newline
Generated QA:\n\{\}\n\newline
(3) Captions:\n\{\}\n\newline
Generated QA:\n

\noindent\rule{\linewidth}{0.4pt}

{\textcolor{blue}{Causal reasoning}}

You are an AI visual assistant that can analyze a single video. You receive a few sentences, each describing the same video you are observing.
The task is to use the provided caption, create a plausible question about the video, and provide the answer in detail.\newline
\textbf{Create questions that explain actions, either uncovering the intentions of the previously occurring actions or stating causes for subsequent actions.}\newline
To answer such questions, one should require first understanding the visual content, then based on the background knowledge or reasoning, either explain why the things are happening that way, or provide guides and help to user's request.  Make the question challenging by not including the visual content details in the question so that the user needs to reason about that first.\newline
Always answer as if you are directly looking at the video.\newline
(1) Captions:\n\{\}\n\newline
Generated QA:\n\{\}\n\newline
(2) Captions:\n\{\}\n\newline
Generated QA:\n\{\}\n\newline
(3) Captions:\n\{\}\n\newline
Generated QA:\n

\noindent\rule{\linewidth}{0.4pt}

{\textcolor{blue}{Short QAs}}

You are an AI visual assistant that can analyze a single video. You receive a few sentences, each describing the same video you are observing.
The task is to use the provided caption, create a plausible question about the video, and \textbf{provide a short answer with less than three words}.\newline
To answer such questions, one should require first understanding the visual content, then based on the background knowledge or reasoning, either explain why the things are happening that way, or provide guides and help to user's request.  Make the question challenging by not including the visual content details in the question so that the user needs to reason about that first.\newline
Always answer as if you are directly looking at the video.\newline
(1) Captions:\n\{\}\n\newline
Generated QA:\n\{\}\n\newline
(2) Captions:\n\{\}\n\newline
Generated QA:\n\{\}\n\newline
(3) Captions:\n\{\}\n\newline
Generated QA:\n

\end{mdframed}
\vspace{-10pt}
\caption{
\textbf{The prompt template to create instruction-following data for temporal reasoning, causal reasoning, and short QAs.}
}
\label{table:supp:instruction_prompts}
\end{table*}

\begin{table*}[!tb]
\begin{mdframed}[align=center,linewidth=1pt]

\textbf{Captions:}
A baby girl on the left side wearing a grey t-shirt is carrying an egg then she throws the egg at the head of the man, then the egg falls on the ground and it breaks on a grey surface.\newline
A man wearing a red t-shirt sitting on his knees is talking with the baby girl on a grey surface.
In the background, there is a grey car, a grey surface, a brown mat, and people speaking and crying sounds are audible.

\textbf{Question:}
What did the baby girl on the left side wearing a grey t-shirt do with the egg after she is carrying it?

\textbf{Answer:}
The girl throws the egg at the head of the man.

\noindent\rule{\linewidth}{0.4pt}

\textbf{Captions:}
A boy wearing black shorts is standing on the side of the swimming pool over small rocks and then he performs a backflip and injured himself.
In the background, there is a swimming pool, rocks, trees, and people's voices and water splashing sound is audible.

\textbf{Question:}
What was the boy wearing black shorts doing before he performing a backflip?

\textbf{Answer:}
The boy is standing on the side of the swimming pool over small rocks. Then he performs a backflip and injured himself.

\noindent\rule{\linewidth}{0.4pt}

\textbf{Captions:}
A boy wearing a black t-shirt rides a black bicycle in a backward direction and falls on a gray surface.
A girl wearing a black cloth is moving on a gray surface, stops, and looks back toward the boy.
In the background, there are gray surfaces, buildings, metallic barrier poles, trees, boats, and the sky is visible, and sounds of people speaking and the wind are audible.

\textbf{Question:}
What did the girl do after the boy wearing a black t-shirt rides a black bicycle in a backward direction and falls on a gray surface?

\textbf{Answer:}
The girl wearing a black cloth is moving on a gray surface. After the boy falls on a gray surface, she stops and looks back toward the boy.

\end{mdframed}
\vspace{-10pt}
\caption{
\textbf{Few-shot examples for prompting temporal-reasoning question-answer pairs.}
}
\label{table:supp:instruction_temporal_examples}
\end{table*}

\begin{table*}[!tb]
\begin{mdframed}[align=center,linewidth=1pt]

\textbf{Captions:}
A baby girl on the left side wearing a grey t-shirt is carrying an egg then she throws the egg at the head of the man, then the egg falls on the ground and it breaks on a grey surface.
A man wearing a red t-shirt sitting on his knees is talking with the baby girl on a grey surface.
In the background, there is a grey car, a grey surface, a brown mat, and people speaking and crying sounds are audible.

\textbf{Question:}
Why did the egg fall on the ground and breaks?

\textbf{Answer:}
The egg was thrown by the girl at the head of the man sitting on his knees. However, it did not hit the man. Therefore, the egg falls on the ground and breaks on a grey surface.

\noindent\rule{\linewidth}{0.4pt}

\textbf{Captions:}
A boy wearing black shorts is standing on the side of the swimming pool over small rocks and then he performs a backflip and injured himself.
In the background, there is a swimming pool, rocks, trees, and people's voices and water splashing sound is audible.

\textbf{Question:}
How did the boy standing on the side of the swimming pool over small rocks injure himself?

\textbf{Answer:}
The boy standing on the side of the swimming pool over small rocks perform a backflip. Instead of falling into the water, he hit on the small rocks. That is why he injured himself.

\noindent\rule{\linewidth}{0.4pt}

\textbf{Captions:}
A boy wearing a black t-shirt rides a black bicycle in a backward direction and falls on a gray surface.
A girl wearing a black cloth is moving on a gray surface, stops, and looks back toward the boy.
In the background, there are gray surfaces, buildings, metallic barrier poles, trees, boats, and the sky is visible, and sounds of people speaking and the wind are audible.

\textbf{Question:}
How did the boy wearing a black t-shirt fall on a gray surface?

\textbf{Answer:}
The boy wearing a black t-shirt rides a black bicycle in a backward direction. Riding a black bicycle backward is slow and hard to balance. Also, the boy is not good at riding a black bicycle backward. As a result, he fell on a gray surface.

\end{mdframed}
\vspace{-10pt}
\caption{
\textbf{Few-shot examples for prompting causal-reasoning question-answer pairs.}
}
\label{table:supp:instruction_causal_examples}
\end{table*}

\begin{table*}[!tb]
\begin{mdframed}[align=center,linewidth=1pt]

\textbf{Captions:}
A baby girl on the left side wearing a grey t-shirt is carrying an egg then she throws the egg at the head of the man, then the egg falls on the ground and it breaks on a grey surface.
A man wearing a red t-shirt sitting on his knees is talking with the baby girl on a grey surface.
In the background, there is a grey car, a grey surface, a brown mat, and people speaking and crying sounds are audible.

\textbf{Question:}
who throws the egg at the man

\textbf{Answer:}
baby girl

\noindent\rule{\linewidth}{0.4pt}

\textbf{Captions:}
A boy wearing black shorts is standing on the side of the swimming pool over small rocks and then he performs a backflip and injured himself.
In the background, there is a swimming pool, rocks, trees, and people's voices and water splashing sound is audible.

\textbf{Question:}
what kind of pool is in the background

\textbf{Answer:}
swimming

\noindent\rule{\linewidth}{0.4pt}

\textbf{Captions:}
A boy wearing a black t-shirt rides a black bicycle in a backward direction and falls on a gray surface.
A girl wearing a black cloth is moving on a gray surface, stops, and looks back toward the boy.
In the background, there are gray surfaces, buildings, metallic barrier poles, trees, boats, and the sky is visible, and sounds of people speaking and the wind are audible.

\textbf{Question:}
what happens when the man loses control

\textbf{Answer:}
falls down

\end{mdframed}
\vspace{-10pt}
\caption{
\textbf{Few-shot examples for prompting short question-answer pairs.}
}
\label{table:supp:instruction_short_examples}
\end{table*}

\end{document}